\documentclass[10pt,twocolumn,letterpaper]{article}

\usepackage{cvpr}
\usepackage{times}
\usepackage{epsfig}
\usepackage{graphicx}
\usepackage{amsmath}
\usepackage{amssymb}
\usepackage{booktabs}
\usepackage{multirow}

\usepackage[pagebackref=true,breaklinks=true,letterpaper=true,colorlinks,bookmarks=false]{hyperref}

\begin{document}

\title{Detecting Morphing Attacks via Continual Incremental Training}

\author{Lorenzo Pellegrini, Guido Borghi, Annalisa Franco, Davide Maltoni\\
Dipartimento di Informatica - Scienza e Ingegneria\\
University of Bologna, 47521 Cesena, Italy\\
{\tt\small \{l.pellegrini, name.surname\}@unibo.it}
}

\maketitle
\thispagestyle{empty}

\begin{abstract}
Scenarios in which restrictions in data transfer and storage limit the possibility to compose a single dataset -- also exploiting different data sources -- to perform a batch-based training procedure, make the development of robust models particularly challenging.
We hypothesize that the recent Continual Learning (CL) paradigm may represent an effective solution to enable incremental training, even through multiple sites.
Indeed, a basic assumption of CL is that once a model has been trained, old data can no longer be used in successive training iterations and in principle can be deleted. 
Therefore, in this paper, we investigate the performance of different Continual Learning methods in this scenario, simulating a learning model that is updated every time a new chunk of data, even of variable size, is available. 
Experimental results reveal that a particular CL method, namely Learning without Forgetting (LwF), is one of the best-performing algorithms. Then, we investigate its usage and parametrization in Morphing Attack Detection and Object Classification tasks, specifically with respect to the amount of new training data that became available.
\end{abstract}

\section{Introduction} \label{sec:intro}
\begin{figure}
    \centering
    \includegraphics[width=1\columnwidth]{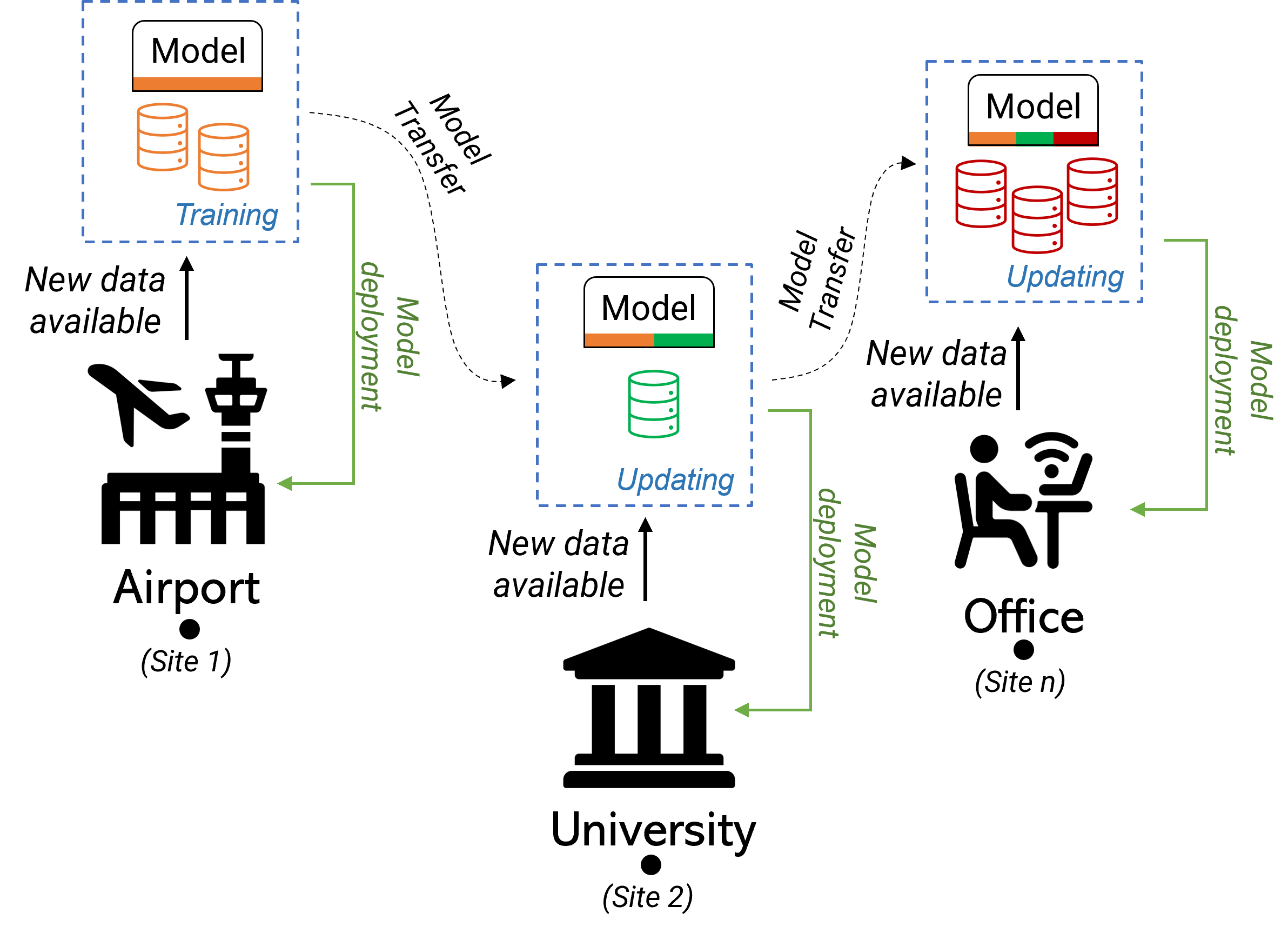}
    \caption{Visualization of the proposed incremental training scenario, in which a trained model is updated every time new chunks of data are temporarily available, even on different sites.
    In this manner, no data transfer is involved and Continual Learning algorithms are eligible to fully exploit the data available through incremental model training.}
    \label{fig:initial}
\end{figure}

In this paper, we address the scenario in which new sets of biometric training data become progressively available across time, even on different sites~\cite{madhavan2021incremental}.
Differently from the traditional Machine Learning setting, the batch-based training procedure~\cite{bisong2019batch} is unfeasible, making challenging the learning process. 
Therefore, we investigate the use of the Continual Learning (CL)~\cite{PARISI201954} paradigm to train a model in a distributed setting, in which several distinct data chunks containing personal information cannot be stored and then are available only in a limited time frame.
In other words, we aim to address the problem of incrementally training a model on multiple data sources that, for different reasons (\eg privacy issues), cannot be shared and stored for long time ranges, thus making it impossible to create a single training dataset, as represented in Figure \ref{fig:initial}.

\begin{figure}
     \centering
     \begin{subfigure}[th!]{0.32\columnwidth}
         \centering
         \includegraphics[width=1\columnwidth]{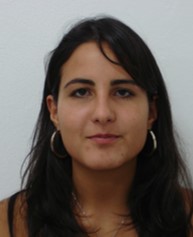}
         \caption{Accomplice}
         \label{fig:accomplice}
     \end{subfigure}
     \hfill
     \begin{subfigure}[th!]{0.32\columnwidth}
         \centering
         \includegraphics[width=1\columnwidth]{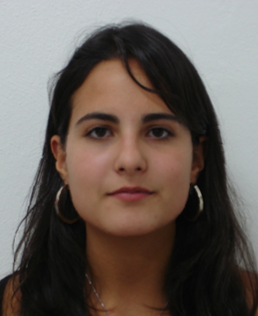}
         \caption{Morphed}
         \label{fig:morphed}
     \end{subfigure}
     \hfill
     \begin{subfigure}[th!]{0.32\columnwidth}
         \centering
         \includegraphics[width=1\columnwidth]{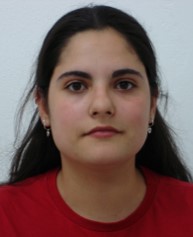}
         \caption{Criminal}
         \label{fig:morphed}
     \end{subfigure}
     \hfill
    \caption{Example of the Morphing Attack~\cite{FerraraFM14}, in which a morphed image is obtained merging the identities of an accomplice and a criminal. The resulting face can fool automatic and human face verification-based controls.
    }
    \label{fig:morphing_attack}
\end{figure}

A practical case is represented by the development of solutions to contrast the \textit{Morphing Attack}~\cite{FerraraFM14} (see Fig.~\ref{fig:morphing_attack}), in which severe privacy issues strongly limit the possibility of storing, transferring and sharing public datasets of sensitive data (\eg facial images, sex, and age). 
As a consequence, each research laboratory or institution usually exploits for training only its own data, thus developing a model with limited generalization capability whose performance is generally unsatisfactory on new unseen data~\cite{borghi2022incremental}.

From a theoretical point of view, this distributed training setting can be tackled through Federated Learning~\cite{li2020federated}, which is indeed a paradigm focused on training a global model exploiting multiple clients that have access to private and not-shareable data. 
We observe that this recent paradigm is a promising solution, even though it presents technical challenges~\cite{konevcny2016federated}, including the development of an infrastructure that supports repeated global model transfers and that maintains a copy of the original dataset in case new training from scratch or fine-tuning procedures of the original model become necessary. 
Moreover, privacy issues can limit the temporal range in which new data are stored on a site, hampering the possibility to have different datasets available on different sites.
For example an airport gate could store some face images only for the short time required to update a model. 
Finally, a limited latency between clients, that have to be simultaneously online, despite the relevant size of data transfers, is needed~\cite{mcmahan2017communication}.
These issues lead us to explore a complementary approach based on the recent and interesting Continual Learning paradigm.

In particular, this paper represents one of the first attempts to investigate the use of this paradigm in the aforementioned scenario, that imposes three challenging and novel peculiarities: 

\begin{itemize}
    
    \item \textbf{Variable chunk size}: data amounts available at each training step (from now referred to as \textit{experience}) are not known in advance. In other words, it is possible to update the model only through a variable amount of training samples. 
    In our \textit{Morphing Attack Detection} (MAD)~\cite{raja2020morphing} scenario, the model is kept updated, for instance, every time a certain (and variable) amount of new data becomes available (\eg a laboratory has collected a new dataset, new morphed images are generated with new algorithms or additional bona fide images are collected through an Automatic Border Control system in an international airport, etc.) 

    \item \textbf{Variable amount of training steps}: the number of learning experiences is not known in advance. It is unpredictable to define how many times a model receives new data to update the knowledge. It follows that we aim to obtain the best performance after each training phase of the model, in order to improve or, at least, fully preserve the model performance in the MAD task. Therefore, a metric able to consider not only the final performance but the accuracy across the whole learning process, referred to as BRoT, is introduced in this paper, as detailed in Section~\ref{sec:brot}.

    \item \textbf{Limits in storage}: limitations in the release and transfer of datasets inhibit the direct use of stored samples. However, we observe that in the CL paradigm, the use of these samples, referred as replay memory~\cite{buzzega2021rethinking}, is one of the most effective ways to contrast the so-called \textit{catastrophic forgetting} problem~\cite{mccloskey1989catastrophic}, \ie the tendency of a model to abruptly and drastically forget the previously learned knowledge. 
    A workaround may consist in exploiting a replay memory based on embeddings instead of real samples~\cite{pellegrini2020latent}, but further investigations related to privacy constraints are still needed and are out of the scope of this paper.
    
\end{itemize}

We observe that, even in the literature related to Continual Learning, these aspects are not yet fully investigated.
Therefore, in this paper, we simulate the proposed incremental training scenario addressing the MAD task. In addition, the validity of our findings is further assessed on the object classification task. 
We test and compare different traditional Continual Learning approaches, focusing in particular on the Learning without Forgetting (LwF)~\cite{li2017learning} method. 

Experimental results reveal that the proposed setting represents an interesting and challenging scenario for common CL methods. Moreover, the choice of the proper parametrization of LwF with variable experience size is not trivial and must be investigated.

\section{Morphing Attack Detection}
\label{sec:mad}
In recent years, several studies~\cite{raja2020morphing, scherhag2019face} confirmed that existing Face Recognition Systems (FRSs) or, more in general, face verification-based algorithms~\cite{borghi2018face}, 
are highly sensitive to specific kinds of manipulation and, in particular, to the morphing process. This vulnerability increases the probability of success of a possible morphing attack, which consists in merging two facial identities (a \textit{criminal} and an \textit{accomplice}) into a single-face morphed image (see Fig.~\ref{fig:morphing_attack}), creating a new hybrid identity and thus destroying the link between the document and its real owner. 
As a consequence, it is possible to deceive the  authority into issuing a document that contains the morphed image. This creates a situation where two individuals can share the same legal document, such as the electronic Machine Readable Travel Document (eMRTD). Once the morphed document is in possession, it can be used to deceive both human officers and automatic face recognition-based controls commonly used, for instance, in airports~\cite{scherhag2017biometric}.

Due to the importance of this type of attack, MAD methods are strongly needed by private and public institutions.
Unfortunately, these methods usually suffer from limited generalization capabilities mainly due to the lack of public datasets, which also hamper the reproducibility of the training procedure. 

These limitations are amplified with MAD models based on deep learning architectures, that are prone to overfit on small low-varied datasets~\cite{borghi2022incremental}.

In this paper, we focus on the specific task of \textit{Differential Morphing Attack Detection} (D-MAD) systems, \ie methods that receive a pair of images as input~\cite{borghi2021double}. In particular, the first image is the one stored in the document (\ie suspected morph) while the second is a trusted live captured image.
These methods work under the assumption that it is possible to compare the two input images (one of which is surely genuine) to detect the presence of the morphing attack. Generally, this approach achieves better performance with respect to MAD systems that receive as input only a single image (S-MAD).
From a general point of view, our MAD task can be considered a binary classification task, with the two classes ``morphed'' and ``bona fide''.

\section{Incremental MAD} \label{sec:probdef}
In this Section, we define the tasks involved in the development of MAD systems incrementally trained. 
In particular, we formulate the terminology used in the rest of the paper and, for the sake of readability, we briefly recall the Continual Learning paradigm.

\subsection{Incremental Training}
Following~\cite{borghi2022incremental}, we formally define two key elements of the proposed scenario:
\begin{itemize}
    \item \textbf{Learning Experience} ($l$): the given model $M$ is trained on a specific chunk of data of variable size. Then, a learning experience is defined as:
    \begin{equation}
        l_i = (M_k, \, d_i), \,  1 \leq i, k \leq N 
    \end{equation}
    where $M_k$ is the model trained at the $k$-th experience and updated using a new set of data $d_i \in D$, where $D=\{d_i, \, i=1,..,N\}$ is the entire set of training data available and $N = |D|$ is the total number of data chunks accessible for the training experiences.
    \item \textbf{Testing experience} ($t$): the given model $M$ is tested after each learning experience on the same set of testing datasets, in order to globally monitor the model performance. Formally:
    \begin{equation}
        t_i = (M_k, \,  E), \,  1 \leq i, k \leq N 
    \end{equation}
    where $M_k$ is the model updated at the $k$-th learning experience, $E$ is the set of the testing datasets and $N = |D|$ is the total number of datasets as before. We observe that the size, the order, and the amount of data chunks are irrelevant to the testing procedure since no training steps are performed. $E$ is a fixed set in order to compute comparable performance metrics after a given training experience. 
\end{itemize}
Therefore, the proposed incremental scenario is formally described as:
\begin{equation}
    B = (l_i, \, t_i), \, i=1,...,N
\end{equation}
or rather as an ordered set of training experiences $l_i \in L$ computed on a specific chunk of data of variable size, each of them followed by a testing experience $t_i \in T$ used to monitor the model performance across time. Since the single chunks of data $d_i$ are not shareable, model $M$ is transferred each time to be updated through Continual Learning techniques to contrast the catastrophic forgetting~\cite{mccloskey1989catastrophic}.

\subsection{Continual Learning}
Continual Learning, also known as lifelong learning, is the ability to continually acquire, fine-tune and transfer knowledge across time~\cite{PARISI201954}. 
This is an ability naturally present in humans and animals, but not in artificial learning systems, especially if based on deep learning architectures. In particular, computational systems, that commonly are trained on stationary batches of training data, have difficulties in acquiring new incremental knowledge from non-stationary data distributions due to the catastrophic forgetting problem~\cite{mccloskey1989catastrophic}.

The Continual Learning paradigm greatly differs from the Machine Learning one, in which the development of a learning agent is divided into two distinct phases: learning and deployment. Indeed, training data, collected only before the learning phase, are unrealistically supposed to be representative of all the nuances of future test data~\cite{graffieti2022continual}.

A variety of approaches have been proposed in the literature to limit or contrast the forgetting, ranging from regularization methods~\cite{ahn2019uncertainty}, that exploit constraints on the update of the neural weights, to dynamic architectures~\cite{douillard2022dytox}, in which changes in architectures are introduced to deal with the new information, and memory replay methods~\cite{van2020brain}, based on the storing of past data used for current training procedures. A further analysis of the CL method investigated in this paper is reported in Section~\ref{sec:baselines}.

\section{Experiments}
\subsection{Datasets}
\noindent \textbf{Idiap Morph}~\cite{sarkar2020vulnerability,sarkar2022gan} collects images belonging to different datasets, \ie  FRGC~\cite{phillips2005overview}, Feret~\cite{phillips1998feret} and Face Research Lab London Set (FRLL)~\cite{debruine2017face}. Morphed images are produced through $5$ different morphing algorithms, \ie OpenCV~\cite{opencv_morph}, FaceMorpher~\cite{facemorpher}, AMSL~\cite{neubert2018extended}, StyleGAN~\cite{karras2020analyzing} and WebMorph~\cite{debruine2017face}.
The quality of resulting morphed images is usually medium-low since artifacts commonly produced by landmark-based morphing algorithms or GAN generation are visible in the majority of images. No manual or automated retouching is applied.

\noindent \textbf{Progressive Morphing Database}~\cite{ferrara2017face} (PMDB) consists of more than $1000$ morphed images produced using the algorithm described in~\cite{ferrara2017face}, and accomplices and criminals are selected in AR~\cite{martinez1998ar}, FRGC~\cite{phillips2005overview}, and Color Feret~\cite{phillips1998feret} datasets. In total, $280$ subjects ($134$ males and $146$ females) are available.
In morphed images are visible some artifacts, such as ghosts and blurred areas, especially close to the nose, eyes, and mouth.

\noindent \textbf{MorphDB}~\cite{ferrara2017face} dataset is composed of $100$ high-quality morphed images, from an equal number of male and female subjects ($50$). This is one of the few datasets in which images have been manually retouched to hide artefacts produced during the morph operation and then is effective to test the performance of MAD systems. This dataset is not publicly available, but the FVC-onGoing~\cite{dorizzi2009fingerprint} platform offers the possibility to test it as a sequestered dataset.

\subsection{Configuration}
Inspired by the state-of-the-art method~\cite{scherhag2020deep} in the D-MAD scenario, the model $M$ is a Multi-Layer Perceptron (MLP) that processes extracted features.
This MLP receives as input features extracted through the ``ArcFace''~\cite{deng2019arcface} network trained on the merge of VGGFace2~\cite{cao2018vggface2}, CASIA~\cite{yi2014learning}, and MS1MV2~\cite{guo2016ms} datasets\footnote{\url{https://github.com/serengil/deepface}}. 

In all experiments, MLP has the same architecture that consists of $5$ layers that have $512, 250, 125, 64, 2$ neurons, respectively. The activation function is ReLU, while the loss function is the Categorical Cross Entropy (CCE). As for the optimizer, we use SGD with a learning rate of $10^{-2}$ and a momentum of $0.9$. No weight decay is applied.

The training set $D$ consists of (morphing algorithm - dataset): StyleGAN - Feret; OpenCV - FRGC; FaceMorpher - FRLL. The whole MorphDB dataset is used for the test, consisting of pairs with both the criminal and the accomplice. In this manner, we perform a challenging cross-dataset evaluation, limiting the influence of overfitting on the investigated algorithms.
Since $|D| = 4$, we permute all possible training dataset orders ($4! = 24$ orders in total) for MAD experiments.
All the training datasets are split into chunks $d_i \in D$ of variable size, depending on the experimental validation conducted and detailed in the following.

\subsection{Metrics} \label{sec:metrics}
MAD task is evaluated through metrics commonly used in the literature~\cite{raja2020morphing}.
The Bona Fide Presentation Classification Error Rate (BPCER) represents the proportion of bona fide images wrongly classified as morphed:
\begin{equation}
    \textnormal{BPCER} (\tau) = \frac{1}{N} \sum_{i=1}^{N} H(b_i - \tau)
\end{equation}
Attack Presentation Classification Error Rate (APCER) represents the proportion of morphed images wrongly accepted as bona fide:
\begin{equation}
    \textnormal{APCER} (\tau) = 1 - \left[ \frac{1}{M} \sum_{i=1}^{M} H(m_i - \tau) \right] 
\end{equation}
In both definitions, $\tau$ is the score threshold on which $b_i, m_i$, the detection scores, are compared; $H(x) = {1 \,\, \text{if} \, x > 0, 0 \, \text{otherwise}}$ is defined as a step function.
Being error rates, low values are desired.

To summarize metrics across different testing experiences $t_i$, we compute the Area Under the Curve (AUC) metric.
AUC is similar to the Average Mean Class Accuracy (AMCA) metric~\cite{graffieti2022continual}, but it is obtained through the trapezoidal rule and the final value is divided by the number of training experiences.
In the MAD task, the AUC is computed by adding the EER, the error rate for which both BPCER and APCER metrics are equal, and the lowest point in which BPCER with APCER $ \leq 1\%$ (typical working point of face verification-based systems). 

\subsubsection{BRoT Metric} \label{sec:brot}
Finally, following the aforementioned considerations about the measurement of performance during the whole learning procedure, we introduce an additional metric named \textit{Borda Ranking over Time} (BRoT), computed over a set of algorithms $\mathcal{A}$, based on the idea of rewarding the algorithms that perform better at each testing experience. Let $r(a_j,t_i)$ be the ranking of algorithm $a_j\in\mathcal{A}$ at the testing experience $t_i$; ranking is here established according to the $\text{BPCER}_{0.001}$ for MAD. At each $t_i$, Borda count~\cite{lippman2013voting} is applied to score the tested algorithms, \ie a decreasing number of points $p(r(a_j,t_i))$ is assigned to each algorithm based on the corresponding ranking, with $p(i)=|\mathcal{A}|-i$. 

\begin{figure*}[th!]
    \centering
    \includegraphics[width=1\textwidth]{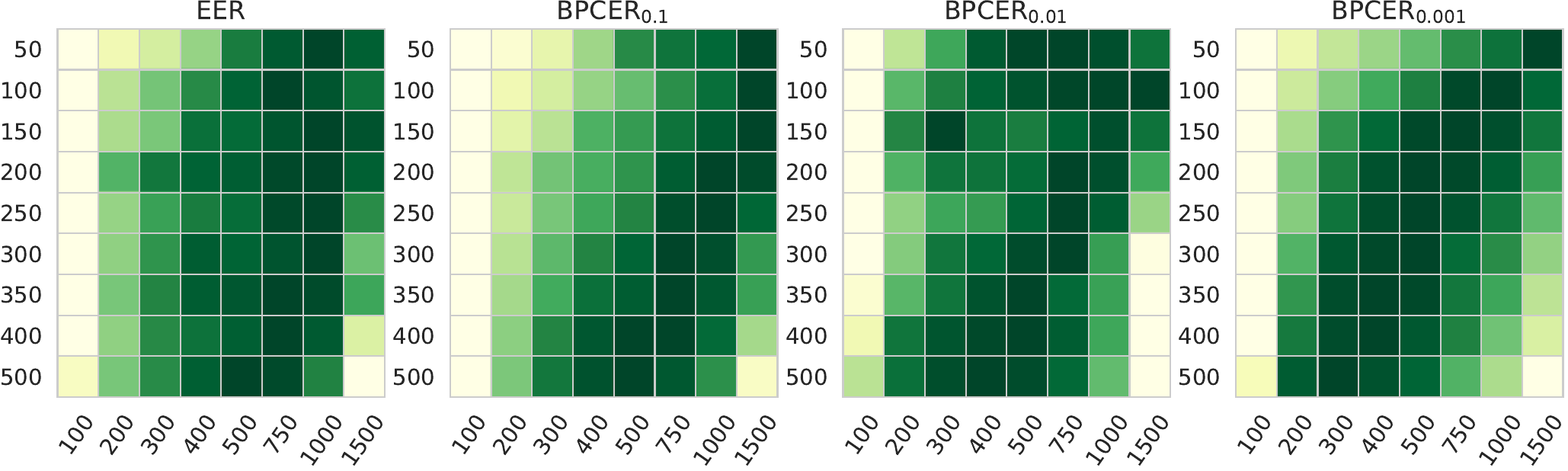}
    \caption{Performance of LwF~\cite{li2017learning} on the MAD task with respect to different training experience sizes ($y$-axis) and $\lambda$ values ($x$-axis). Each matrix is referred to a specific metric commonly used in the MAD scenario, \ie EER, $\text{BPCER}_{0.1}$, $\text{BPCER}_{0.01}$ and $\text{BPCER}_{0.001}$ (the most challenging case), as detailed in Section~\ref{sec:metrics}. The darker color is better. As shown, different $\lambda$ values strongly impact the performance of the LwF algorithm, and suggest that larger values are the best choice for small experience sizes.}
    \label{fig:mat_mad}
\end{figure*}

For each algorithm $a_j$, the points are accumulated over the different learning experiences and the total score is finally normalized by the maximum theoretical score:
\begin{equation}
    \textnormal{BRoT}(a_j) =\frac{ \sum_{i=1}^Np(\mathcal{R}(a_j,t_i))}{|\mathcal{A}|\times N} 
\end{equation}
where $N$ is the total number of testing experiences.

We observe the AUC metric is useful to understand the whole performance of the method across all the training experiences, highlighting the performance in terms of the best accuracy achieved. 
Differently, the BRoT metric enables the understanding of which algorithm has the greatest probability of having high accuracy across the whole learning process (without taking into consideration the possible gap between different methods in terms of absolute accuracy). 
These two metrics are complementary and help to understand different aspects of the performance of the investigated model in our novel scenario.

\section{Results on MAD}

\subsection{Baselines} \label{sec:baselines}
Different Continual Learning methods have been investigated.
Elastic Weight Consolidation (\textbf{EWC}) \cite{kirkpatrick2017overcoming} is based on a penalty loss that tries to constrain the model weights in maintaining the same value in new experiences.
Learning without Forgetting (\textbf{LwF})~\cite{li2017learning} contrasts the forgetting problem by exploiting two different models: the old model $M_{t-1}$, which is the result of training up to the current experience and thus carries the knowledge of previous data, and the current model $M_t$, which is initialized as a copy of the old model. The old model is frozen and is used through distillation to train the current model by adding a distillation component to the loss function. In this algorithm, an important hyperparameter is represented by $\lambda$, which acts as a regularization term used to balance the two components of the loss function. The value of the $\lambda$ hyperparameter determines the trade-off between preserving knowledge from previous experiences and adapting to the new one: higher values of $\lambda$ are used to emphasize the preservation of knowledge from previous experiences, while lower values favor the learning from current data. With $\lambda=0$, no distillation happens and LwF collapses to a plain fine-tuning strategy. In LwF $\lambda$ is fixed from the beginning and does not change between experiences. This choice may be suboptimal and an adaptive choice of $\lambda$ may be beneficial, as demonstrated in the next sections.

\begin{table}[h!]
\centering
\begin{tabular}{rcc}
\textbf{}      & \multicolumn{2}{c}{\textbf{MAD} $\downarrow$}  \\
\textbf{Exp. Size} & \textbf{Small}  & \textbf{Large}       \\
\toprule
\textbf{Naive}                                  & +14\%              & +16\%             \\
\midrule
\textbf{EWC} \cite{kirkpatrick2017overcoming}   & +14\%              & +16\%             \\
\textbf{SI} \cite{zenke2017continual}           & +21\%              & +24\%             \\
\textbf{SLDA} \cite{hayes2019memory}            & \multicolumn{2}{c}{+27\%}           \\
\textbf{LwF} \cite{li2017learning}              & \textbf{+13}\%              & \textbf{+12}\%          \\
\bottomrule \\
\end{tabular}
\caption{Comparison of different CL methods with variable experience sizes. Results are expressed as the percentage variations of the AuC with respect to the ideal case, \ie the Joint approach. A positive value indicates a higher error, lower values are desired.
}
\label{tab:results_mad}
\end{table}

We include in our analysis also the Synaptic Intelligence (\textbf{SI}) \cite{zenke2017continual} method, based on a quadratic regularization
that aims to preserve the weights that contribute to the performance
on old data. 
Finally, we investigate the Deep Streaming Linear Discriminant Analysis (\textbf{SLDA}) \cite{hayes2019memory} method that, taking inspiration from the data mining research field, uses a covariance matrix to perform the final prediction on pre-extracted features. This method, expressively developed for the Online Learning task~\cite{graffieti2022continual}, does not have a proper learning phase (intended as the learning process of common neural networks) and the concept of batch size, since it processes one sample per time in a streaming manner, without any memory mechanism.

\begin{figure*}[th!]
    \centering
    \includegraphics[width=1\textwidth]{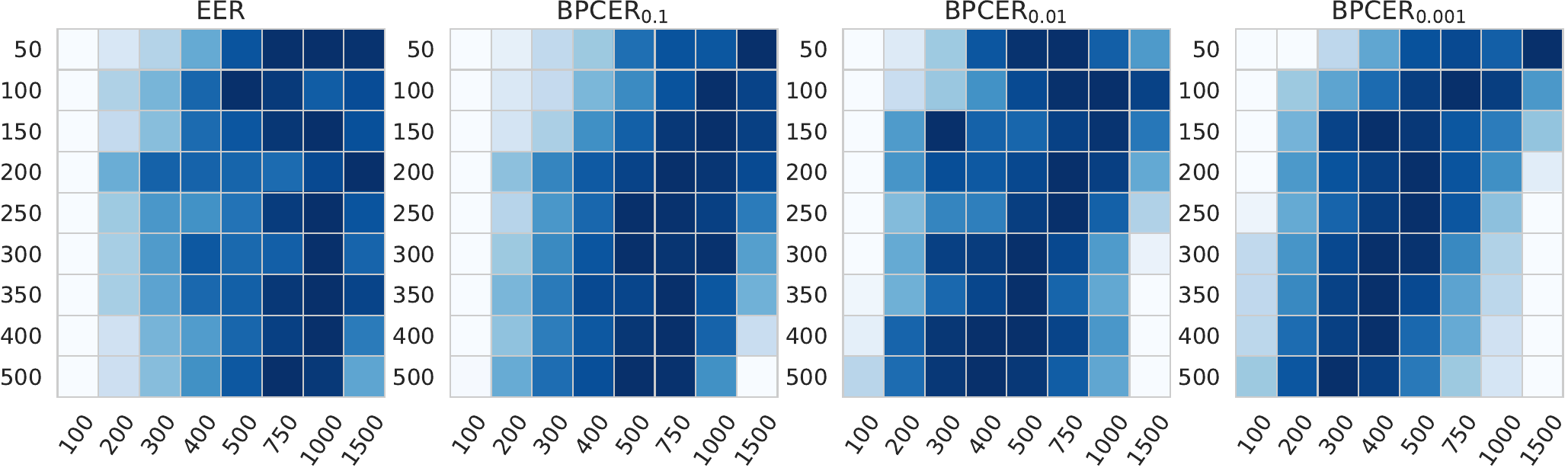}
    \caption{Performance of LwF~\cite{li2017learning} on the MAD task in terms of the BRoT metric (see Sect.~\ref{sec:metrics}), which highlights the probability to have the best performance across the whole distributed training process. The darker color is better.}
    \label{fig:mat_mad_brot}
\end{figure*}

In order to have a reference in results, we also implement two additional approaches: the first is the \textbf{naive} method, in which the model is optimized on available data, without any specific mechanism to contrast the catastrophic forgetting, while the second is the \textbf{Joint} method, in which all the training data are available at the beginning of the training procedure (\ie the common Machine Learning scenario). 
All the compared baselines have been tested by exploiting the public implementations available in Avalanche~\cite{lomonaco2021avalanche}.

\subsection{Results} \label{sec:results}
Firstly, we test baselines in our distributed training scenario with variable experience sizes. Results are reported in Table~\ref{tab:results_mad}, for two different settings referred to as ``\textbf{small}'' and ``\textbf{large}'', respectively; in the first one, the size of experience varies in the range $[50, 500]$, and the probability to sample a specific size is modelled through the \textit{Zipf}'s law~\cite{li2002zipf}. In the second, the range is inverted $[500, 50]$, and the same distribution is exploited.
In other words, in the case ``small'' there is a high probability to have small data chunks (of variable size) as input, and vice versa in the ``large'' one.

In this test, the upper bound is represented by the performance of the Joint training, in which the training set is available as a single chunk in a single site, and then results are expressed as the deviation percentage of the AUC metric (see Sect.~\ref{sec:metrics}) with respect to the one obtained with the Joint approach. 
Since AUC is related to errors and lower deviation values are better, results reveal that the distributed training scenario is challenging and that the size of small and large only partially impact the general performance. As expected, the SLDA method reports the same accuracy since it is not influenced by the experience size (see Sect.~\ref{sec:baselines}).

From a general point of view, the LwF approach tends to achieve the best performance and is therefore selected for the next investigation that is focused on the analysis of the weight assigned to the distillation loss with respect to the experience size.
It is worth noting that this aspect is not yet fully investigated in the literature, especially in relation to different experience sizes.

In particular, we create data chunks with different fixed sizes, starting from $50$ with a step of $50$ ($|d| = \{50, 100, 150, ..., 500\}$).
Then, differently from the previous case, the size of chunks is the same across the whole training procedure.
For each defined size, we test different $\lambda$ values: empirically, we found a specific range must be used to achieve reasonable results, in particular a range of $\{100, 1500\}$ for the MAD task.
Results, obtained by averaging the test metrics on all possible dataset order configurations, are condensed through colored matrices (the darker color is better), as reported in Figures \ref{fig:mat_mad}.
We observe that different $\lambda$ strongly impact the performance of the LwF, especially in relation to different training experience sizes. 
For instance, with $\lambda = 100$, the model performance is generally limited for all possible chunk sizes. 
From a general point of view, we observe a trend in which with small experience sizes it is better to have larger $\lambda$ values, and vice versa. 
This tendency is particularly noticeable with $\text{BPCER}_{0.001}$ (last matrix of Fig.~\ref{fig:mat_mad}), which represents the most challenging (and realistic) case of the MAD task. 

\begin{table*}[th!]
\centering
\begin{tabular}{l rcc cccc}
                      \textbf{Training} & \textbf{Method} & \textbf{Classifier} & \textbf{Exp. Size} & \textbf{EER} & \textbf{$\text{BPCER}_{0.1}$} & \textbf{$\text{BPCER}_{0.01}$} & \textbf{$\text{BPCER}_{0.001}$} \\
                     \toprule
\multirow{2}{*}{Batch} &\textbf{ArcFace}~\cite{scherhag2020deep} & SVM & Joint  & 0.121              & 0.135                   & 0.275                     & 0.528                      \\
 &\textbf{ArcFace}~\cite{scherhag2020deep}  & MLP & Joint  & 0.138         & 0.209             & 0.722               & 0.825                \\
\midrule
\multirow{3}{*}{Incremental}&\textbf{LwF}~\cite{li2017learning}   & MLP & small & 0.156        & 0.294             & 0.837               & 0.837\\
&\textbf{LwF}~\cite{li2017learning}   & MLP & large & 0.145        & 0.230             & 0.786               & 0.922                \\
&\textbf{LwF}~\cite{li2017learning}    & MLP & fixed & 0.140        & 0.224             & 0.679               & 0.896                 \\
\bottomrule \\
\end{tabular}
\caption{Experimental results obtained on the MorphDB dataset for the MAD task. In particular, it is possible to compare the performance of the sota MAD method~\cite{scherhag2020deep}, based on the common batch-based training (first row), with respect to the investigated LwF method for incremental training. In the Joint scenario, all training data are available at the same time, while ``small'', ``large'' and ``fixed'' refers to the incremental training described in Section \ref{sec:results}.}
\label{tab:res_morphing}
\end{table*}

Then, we test LwF in terms of BRoT metric, reporting the results (expressed in the same visual form of colored matrices) in Figure~\ref{fig:mat_mad_brot}.
As shown, also the proposed BroT metric confirms the tendency noted in the previous cases, revealing that the proper choice of $\lambda$ is needed to enhance the probability to achieve the best performance on the whole distributed training procedure.

Finally, we test MAD capabilities of the investigated solutions.
Results are reported in Table~\ref{tab:res_morphing}, in which we show the performance of the following methods, all working on features extracted through the ArcFace architecture~\cite{deng2019arcface} from the two input images, and combined by a subtraction: 
i) the current state-of-the-art method described in~\cite{scherhag2020deep}, consisting in an SVM classifier with the Radial Basis Function kernel. The training scenario is Joint, since all data must be available before the single batch-based training procedure;
ii) a solution equal to the previous one, but exploiting MLP as a classifier. This classifier has been adopted to simplify the comparison with the incremental training methods, in which we are forced to use a deep learning-based architecture to apply the CL strategies;
iii) three incremental training approaches based on the investigated LwF method, trained with different experience sizes: large, small and the scenario with fixed experience that achieved the best performance (experience size equal to $500$ and $\lambda = 200$). 

As expected, the first method based on SVM and batch training exhibits the best performance in terms of EER and BPCER; such results represent in our experiments a sort of upper bound to the performance achievable in this scenario. Batch training, in fact, typically outperforms incremental learning and SVM proved to be the best classifier coupled with ArcFace features for the DMAD task. The MLP classifier unfortunately performs slightly worse, especially at BPCER levels corresponding to a low error threshold.
Interestingly, all the investigated CL strategies have similar performance with respect to the batch-based training: then, we observe that LwF method is a promising method to bridge the gap between the common Machine Learning training scenario and the incremental training one, needed to deal with highly constrained scenarios.

 \begin{table}[th!]
\centering
\begin{tabular}{r cc}
   & \multicolumn{2}{c}{\textbf{Classification} $\uparrow$} \\
\textbf{Exp. Size}  & \textbf{Small}       & \textbf{Large}       \\
\toprule
\textbf{Naive}                                               & -6.4\%                  & -4.8\% \\
\midrule
\textbf{EWC} \cite{kirkpatrick2017overcoming}   & -5.7\%                  & -5.1\% \\
\textbf{SI} \cite{zenke2017continual}           & -9.2\%                  & -5.2\% \\
\textbf{SLDA} \cite{hayes2019memory}            & \multicolumn{2}{c}{-2.7\%} \\
\textbf{LwF} \cite{li2017learning}              & \textbf{-2.2}\%                  & \textbf{-1.5}\% \\
\bottomrule\\
\end{tabular}
\caption{Comparison of different Continual Learning methods with variable experience sizes. Results are expressed in terms of the percentage variations of the AuC with respect to the ideal case, \ie the Joint approach. Note that a negative value indicates a lower accuracy, and then higher values are desired.
}
\label{tab:results_cls}
\end{table}

\section{Further Investigation}
To validate our findings, we extend our investigation also on the supervised continual learning object classification task~\cite{SheOpenLoris, PASQUALE2019260}, which is one of the most common tasks in the Continual Learning field. 
This task consists in continually training a classifier able to incrementally learn new instances, new classes, or both. 
In particular, to maintain similarity with the MAD task, we assume to work in the New Instances (NI) scenario, also referred to as Data Incremental~\cite{delange2021continual}, in which new instances of the same pre-defined classes become progressively available during the training phase.
We observe this scenario is slightly different from the Domain Incremental (Domain-IL)~\cite{van2019three} task, in which new instances belong also to different domains.

In our validation, we use \textbf{CORe50}~\cite{core50} dataset, that contains $50$ objects, belonging to $10$ categories, acquired in $11$ sessions ($8$ indoor and $3$ outdoor) with different backgrounds. The dataset is organized as reported in the original paper, in which $8$ sessions are used for training and validation and the remaining $3$ for the testing procedure. The total amount of frame is about $164$k with a resolution of $128 \times 128$ pixels.
From the classification task point of view, this dataset is challenging due to changes in backgrounds (outdoor and indoor) and light sources, occlusions and low data variability during the same experience.

In order to reproduce the MAD training setting, we use a ResNet-50~\cite{he2016deep} model, trained on the ImageNet~\cite{deng2009imagenet} dataset, to extract features that are then classified by the same MLP architecture used in the MAD task. 
In this case, $|D| = 1$ and then we shuffle the dataset averaging the collected results on $10$ runs. 
As metrics, we exploit the Top-1 accuracy, in which high values are positive. 
AUC and BroT metrics are based on obtained accuracy values across the testing experiences. 

Results are reported in Table~\ref{tab:results_cls}, in which large positive variations are better. We observe that the performance of the investigated CL methods is generally closer to the Joint baseline with respect to MAD task. 
LwF is confirmed as one of the best methods to limit the drop introduced by our distributed scenario, even though SLDA shows better behaviour in comparison to the MAD task.

\begin{figure}
     \centering
     \begin{subfigure}[b]{0.49\columnwidth}
         \centering
         \includegraphics[width=1\columnwidth]{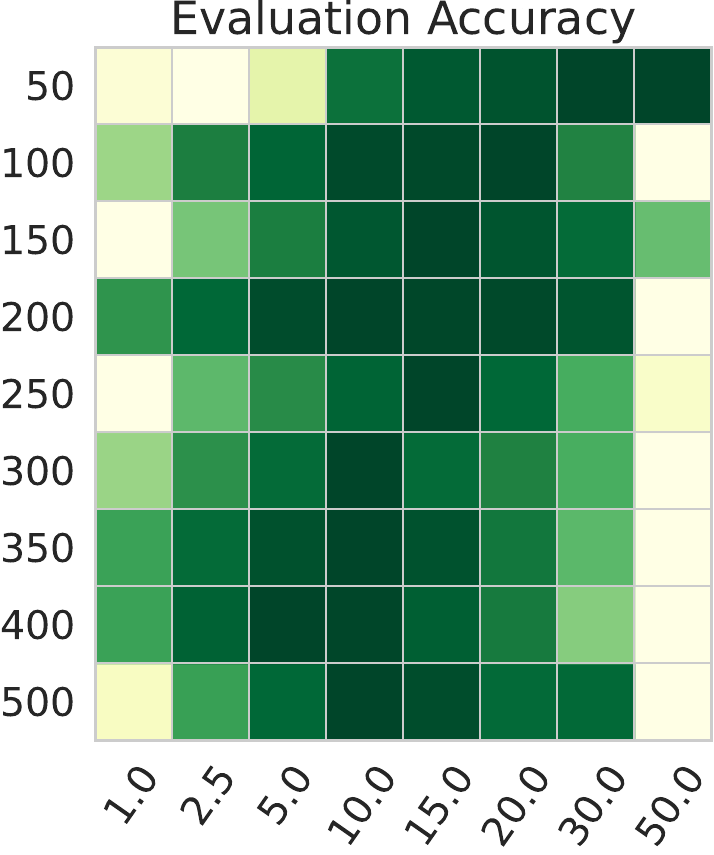}
         \caption{AUC}
         \label{fig:mat_cls}
     \end{subfigure}
     \hfill
     \begin{subfigure}[b]{0.49\columnwidth}
         \centering
         \includegraphics[width=1\columnwidth]{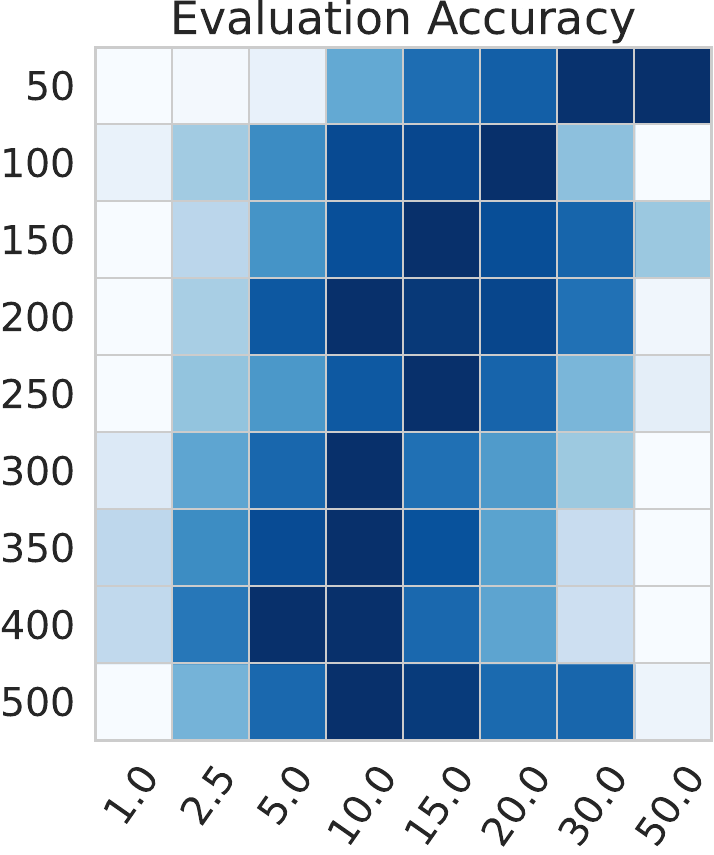}
         \caption{BRoT}
         \label{fig:mat_cls_brot}
     \end{subfigure}
     \hfill
    \caption{Performance of LwF~\cite{li2017learning} on the object classification task with respect to different training experience sizes ($y$-axis) and $\lambda$ values ($x$-axis). The darker color is better. 
    }
    \label{fig:three graphs}
\end{figure}

Focusing our analysis on LwF, we create several chunks of data with a fixed size in order to analyze the impact of varying $\lambda$. Differently from MAD, empirical experiments suggest the proper range to achieve reasonable results is  $\lambda = \{1, 50\}$.
The visualization of the experimental results is reported in Figure~\ref{fig:mat_cls} and \ref{fig:mat_cls_brot} for the AUC and BRoT metrics, respectively. 
Results confirm our previous considerations on MAD task, \ie lambda greatly impacts the performance of LwF, in particular with different sizes, even though this tendency seems to be less evident.
In particular, $\lambda=50$ is a proper choice only for the case in which the chunk size is equal to $50$, while this value leads to a significant drop in performance in all the other cases.

\section{Concluding Remarks}
In this paper, we have carried out one of the first investigations about the performance of Continual Learning methods in an incremental and distributed training scenario addressing the Differential Morphing Attack Detection (D-MAD) task. 
In this scenario, in which data cannot be transferred between different sites due to privacy issues, the recent CL paradigm proves to be useful in enabling model transfer instead of data transfer.

Since data chunks available at each training experience may have different sizes in realistic usage, a further investigation has been conducted to analyze the proper parametrization for the LwF approach, an element not yet fully investigated in the CL literature.
It is worth noting that, from a general point of view, the choice of the distillation loss value ($\lambda$) in the LwF approach is challenging, since it varies in relation to both experience size and task, and a wrong choice can lead to a significant drop in accuracy, as shown in the experimental evaluation.

The outcomes of our analysis can be summarized as follows: 
i) experimental results confirm the opportunity to use the Continual Learning paradigm, and specifically the LwF method, to train a MAD detector in a distributed and incremental manner in order to overcome privacy issues; 
ii) it clearly emerged, in view of future work, the need to automatically determine the proper parametrization of LwF, in terms of the value of $\lambda$ with respect to the size of the training chunk, following the general consideration that small experience size should need larger values;
iii) it is also important to note that further analysis is needed in order to properly determine the $\lambda$ ranges in relation to a specific dataset or task to be addressed;
iv) additional research investigations are important to improve the final accuracy of the MAD model trained in the incremental and distributed setting, that still suffers in terms of performance with respect to a model trained in the common Machine Learning setting (batch training on the whole dataset);.

In conclusion, we believe that our findings can be useful in future research work in the field of Morphing Attack Detection, in order to enable distributed and incremental training, overcome privacy issues and train new models on more varied and large datasets, but also the  Continual Learning task, to properly define values of the $\lambda$ parameter taking into consideration the size of the training experience.

\section*{Acknowledgment}
This work is part of the iMARS project. The project received funding from the European Union’s Horizon 2020 research and innovation program under Grant Agreement No. 883356. Disclaimer: this text reflects only the author’s views, and the Commission is not liable for any use that may be made of the information contained therein.

{\small
\bibliographystyle{ieee}
\bibliography{egbib}
}

\end{document}